\pdfoutput=1

\documentclass[11pt]{article}

\usepackage{ACL2023}

\usepackage{times}
\usepackage{latexsym}

\usepackage[T1]{fontenc}

\usepackage[utf8]{inputenc}

\usepackage{microtype}

\usepackage{inconsolata}

\usepackage{multirow}
\usepackage{makecell}

\usepackage{times}
\usepackage{latexsym}
\usepackage{graphicx}
\usepackage{subfigure}
\usepackage{float}
\graphicspath{{images}}
\usepackage{stfloats}
\usepackage{booktabs}
\usepackage{indentfirst}

\usepackage{amssymb}
\usepackage{amsmath}
\usepackage{amsthm}
\usepackage{url}            

\usepackage{listings}
\usepackage{xcolor}

\usepackage{graphicx}

\usepackage{wrapfig,lipsum,booktabs}

\definecolor{codegreen}{rgb}{0,0.6,0}
\definecolor{codegray}{rgb}{0.5,0.5,0.5}
\definecolor{codepurple}{rgb}{0.58,0,0.82}
\definecolor{backcolour}{rgb}{0.95,0.95,0.92}

\lstdefinestyle{mystyle}{
    backgroundcolor=\color{backcolour},   
    commentstyle=\color{codegreen},
    keywordstyle=\color{magenta},
    numberstyle=\tiny\color{codegray},
    stringstyle=\color{codepurple},
    basicstyle=\ttfamily\small,
    breakatwhitespace=false,         
    breaklines=true,                 
    captionpos=b,                    
    keepspaces=true,                 
    numbers=left,                    
    numbersep=5pt,                  
    showspaces=false,                
    showstringspaces=false,
    showtabs=false,                  
    tabsize=2
}

\usepackage{multirow}

\usepackage{CJKutf8}

\usepackage{amsmath}
\usepackage{amssymb}

\usepackage{algorithm}
\usepackage{algorithmic}

\usepackage{CJKutf8}

\usepackage{CJKutf8}

\title{FT-MDT: Extracting Decision Trees from Medical Texts via a Novel Low-rank Adaptation Method}

\author{
Yuheng Li\textsuperscript{1,$*$}, Jiechao Gao\textsuperscript{2,$\dagger$}\thanks{\ \ These authors contribute equally.}, Wei Han\textsuperscript{3}, Wenwen Ouyang\textsuperscript{4}, \\
\textbf{Wei Zhu\textsuperscript{5,}\thanks{\ \ Corresponding author. For any inquiries, please contact: michaelwzhu91@gmail.com, jiechao@stanford.edu. }, Hui Yi Leong\textsuperscript{6} } \\
\small \textsuperscript{1}Johns Hopkins University, Baltimore, MD, United States \ \ \ \small \textsuperscript{2}Stanford University, Stanford, CA, United States \\
\small \textsuperscript{3}Independent Researcher, Austin, TX, United States. Email: palebluedot.milkyway@gmail.com \\
\small \textsuperscript{4}Carnegie Mellon University, Pittsburgh, PA, United States \ \ \ \small \textsuperscript{5}University of Hong Kong, Hong Kong, HK, China \\
\small \textsuperscript{6}University of Chicage, Chicago, IL, United States \\
}

\begin{document}
\maketitle
\begin{abstract}

Knowledge of the medical decision process, which can be modeled as medical decision trees (MDTs), is critical to building clinical decision support systems. However, current MDT construction methods rely heavily on time-consuming and laborious manual annotation. To address this challenge, we propose PI-LoRA (Path-Integrated LoRA), a novel low-rank adaptation method for automatically extracting MDTs from clinical guidelines and textbooks. We integrate gradient path information to capture synergistic effects between different modules, enabling more effective and reliable rank allocation. This framework ensures that the most critical modules receive appropriate rank allocations while less important ones are pruned, resulting in a more efficient and accurate model for extracting medical decision trees from clinical texts. Extensive experiments on medical guideline datasets demonstrate that our PI-LoRA method significantly outperforms existing parameter-efficient fine-tuning approaches for the Text2MDT task, achieving better accuracy with substantially reduced model complexity. The proposed method achieves state-of-the-art results while maintaining a lightweight architecture, making it particularly suitable for clinical decision support systems where computational resources may be limited.

\end{abstract}

\begin{CJK*}{UTF8}{gbsn}

\section{Introduction}
\label{sec:introduction}

Medical decision processes are critical to building effective clinical decision support systems, and these processes can be effectively modeled as medical decision trees (MDTs). The ability to automatically extract MDTs from clinical guidelines and textbooks would significantly reduce the reliance on time-consuming manual annotation while enabling the development of more robust decision support tools. However, the complex hierarchical nature of medical decision-making presents unique challenges for automated extraction, requiring sophisticated natural language processing techniques that can capture both the structural and semantic aspects of clinical guidelines.

Previous research \cite{zhu2024text2mdt} on the Text2MDT task has primarily focused on either end-to-end frameworks using large language models or pipeline approaches, without fully investigating parameter-efficient fine-tuning (PEFT) methods \cite{Ding2022DeltaTA,qin2023chatgpt,PromptCBLUE,text2dt_shared_task,Text2dt,zhu_etal_2021_paht,Li2023UnifiedDR,Zhu2023BADGESU,Zhang2023LECOIE,Zhu2023OverviewOT,guo-etal-2021-global,zhu-etal-2021-discovering,Zheng2023CandidateSF,info:doi/10.2196/17653,Zhang2023NAGNERAU,Zhang2023FastNERSU,Wang2023MultitaskEL,Zhu2019TheDS,Zhu2021LeeBERTLE,Zhang2021AutomaticSN,Wang2020MiningIH} for this specific application. In addition, while PEFT methods like LoRA \cite{hu2021lora,huang2023c,li2023cmmlu,Cui2023UltraFeedbackBL,wang2024ts,yue2023-TCMEB,Zhang2023LearnedAA,2023arXiv230318223Z,Xu2023ParameterEfficientFM,Ding2022DeltaTA,Xin2024ParameterEfficientFF} have shown promise in reducing model complexity, existing LoRA-based approaches for Text2MDT suffer from critical limitations in rank allocation. Methods such as AdaLoRA \cite{Zhang2023AdaptiveBA} rely on sensitivity-based importance scores that are unreliable as they only consider how a single parameter change affects the model under the assumption that no other parameters change. Similarly, approaches like SoRA \cite{Ding2023SparseLA} and SaLoRA \cite{Hu2023StructureAwareLA} use architectural parameters that cannot reliably reflect the quality or importance of LoRA ranks, leading to suboptimal performance in the context of medical decision tree extraction where precise structural representation is essential.

To address these limitations, we propose PI-LoRA (Path-Integrated LoRA), a novel low-rank adaptation method that overcomes the deficiencies of existing rank allocation techniques for the Text2MDT task. Our approach draws inspiration from Shapley value theory, treating each LoRA module as an independent participant in a cooperative game to measure its contribution to overall model performance. Additionally, we integrate gradient path information to capture synergistic effects between different modules, enabling more effective and reliable rank allocation. This framework ensures that the most critical modules receive appropriate rank allocations while less important ones are pruned, resulting in a more efficient and accurate model for extracting medical decision trees from clinical texts.

Extensive experiments on medical guideline datasets demonstrate that our PI-LoRA method significantly outperforms existing PEFT approaches for the Text2MDT task, achieving better accuracy with substantially reduced model complexity. The proposed method achieves state-of-the-art results while maintaining a lightweight architecture, making it particularly suitable for clinical decision support systems where computational resources may be limited. Our contributions include the first comprehensive exploration of PEFT methods for Text2MDT, a novel PI-LoRA framework that overcomes limitations in existing rank allocation methods, and empirical evidence showing that our approach achieves superior performance compared to both pipeline and end-to-end LLM-based methods while requiring significantly fewer parameters.

\section{Related works}
\label{sec:related_works}

\subsection{Tree data extraction from text}

There is a rich history of NLP tasks that aim to extract tree structures from a given text. The most fundamental task in NLP is syntax analysis, which aims to express the syntactic structure of a sentence into a syntactic tree \cite{Zhang2020ASO,zheng2024sca,zhang2024milora,zeng2025janusvlndecouplingsemanticsspatiality,zeng2025FSDrive,lu2022understanding,wang2025target,wang2024scantd,niu2025decoding,zhang2023moqagpt,wang2024coreinfer,wang2025anglesdontlieunlocking,liu2025qfft,liu2024rag,wang2025reasoningenhanceddomainadaptivepretrainingmultimodal,Leong2025AMASAD}. Parsing often relies on a specific grammar, which is used to refine the output structures of syntax and semantics. Two of the most popular grammars are constituent parsing and dependency parsing. Text2Tree is also seen in many application scenarios. Math word problems (MWPs) \cite{Zhang2022MultiViewRC,Zhao2023AutomaticMS} extract mathematical expressions from the unstructured texts and try to improve the neural networks' capabilities in math problem solving by asking the model to understand the tree structure. Semantic parsing \cite{Kamath2018ASO}, which transforms unstructured text into an SQL query, has promising application potential in areas like dialogue systems, search engines, and business intelligence. Text2MDT \cite{zhu2024text2mdt,text2dt_article,Text2dt,text2dt_shared_task} aims to automatically extract medical decision trees from clinical guidelines and textbooks to support the development of clinical decision support systems without relying on time-consuming manual annotation. It proposes two different approaches for Text2MDT - an end-to-end framework using large language models and a pipeline framework - demonstrating that the LLM-based method outperforms the pipeline approach while a lightweight pipelined method achieves comparable performance with significantly smaller model complexity.

\subsection{Parameter efficient fine-tuning}

Parameter-efficient fine-tuning (PEFT) optimizes only a small subset of new parameters while freezing the backbone model during LLM adaptation \citep{Ding2022DeltaTA,Zhang2023LearnedAA,zhu2024iapt,zhu-tan-2023-spt,Liu2022FewShotPF,xie2024pedro,Cui2023UltraFeedbackBL,zheng2024nat4at,zhu2023acf,gao2023f,zuo-etal-2022-continually,zhang-etal-2022-pcee,sun-etal-2022-simple,zhu-etal-2021-gaml,Zhu2021MVPBERTMP,li-etal-2019-pingan,zhu2019panlp,zhu2019dr,zhou2019analysis,zhang2025time,wang2025ts,liu2025parameter,yi2024drum,tian2024fanlora}. Contemporary approaches fall into three categories: (a) \textit{Addition-based methods} that incorporate supplementary modules (e.g., Adapters \citep{houlsby2019parameter}, Prefix/Prompt tuning \citep{li2021prefix,lester2021power}); (b) \textit{Specification-based methods} that selectively adjust or prune internal parameters \citep{BenZaken2021BitFitSP}; and (c) \textit{Reparameterization-based methods} that project adaptive parameters into low-dimensional subspaces \citep{aghajanyan-etal-2021-intrinsic}, building on the intrinsic dimensionality concept.

LoRA \citep{hu2021lora} exemplifies reparameterization by optimizing low-rank decompositions of weight updates, showing strong performance across models \citep{2023arXiv230514314D}. However, its fixed-rank design lacks guidance for module-specific rank allocation. Subsequent methods address this by dynamically adapting LoRA parameters: (a) DyLoRA \citep{valipour2023dylora} trains multi-rank modules simultaneously via random rank sampling. (b) AdaLoRA \citep{Zhang2023AdaptiveBA} allocates weights using SVD ($\Delta W = P\Lambda Q$) and importance-based pruning. (c) SoRA \citep{Ding2023SparseLA} reduces redundancy via $l_0$ regularization and proximal gradients. (d) SaLoRA \citep{Hu2023StructureAwareLA} enables module-differentiated ranks using Lagrange multipliers.

\section{Method}

\subsection{Task formulation }

The Text2MDT task involves the systematic reconstruction of medical decision trees from clinical documentation. Given a textual input $X= [x_1,\,x_2,\,......,\,x_{n_{text}}]$ comprising $n_{text}$ lexical units, the objective is to synthesize the pre-order traversal sequence $T= [N_1,\,N_2,\,......,\,N_{n_{node}}]$ representing $n_{node}$ structural elements within the medical decision tree. This encoding uniquely captures the hierarchical decision logic embedded in clinical guidelines.

Each decision node integrates three functional components as formalized below:
\begin{align}
\text{Node} & =\{\text{Role}, \, \text{Triplets}, \, \text{Logical\_Rel} \}, \nonumber\\
\text{Role} & = \Diamond \text{ or } \Box, \nonumber\\
\text{Triplets} & = (t_1,t_2, ..., t_{n_{tri}}), \nonumber\\
\text{Logical\_Rel} & = \text{and}, \ \text{or}, \ \text{null},\label{eq:node_structure}
\end{align}
with critical specifications: (a) $\text{Role}$ distinguishes node functionality—$\Diamond$ indicates conditional assessment nodes, whereas $\Box$ designates therapeutic decision nodes. (b) $\text{Triplets}$ aggregates $n_{tri}$ subject-relation-object constructs $(t_1,t_2, ..., t_{n_{tri}})$, each $t=(sub,\,rel,\, obj)$ encoding clinical entities or procedural instructions. (c) $\text{Logical\_Rel}$ establishes inter-triplet connectivity, defaulting to null when $n_{tri} \leq 1$.

\subsection{Prompt template}

According to \cite{zhu2024text2mdt}, the Text2MDT task contains three sub-tasks: (a) triplet extraction; (b) node grouping; (c) tree assembling. The prompt templates and introductions to these sub-tasks are presented in Appendix \ref{sec:appendix_prompt_templates}.

We also consider utilizing the LLMs for the end-to-end framework. Since this task is complex, it is natural that the idea of chain-of-thought (COT) \cite{Wei2022ChainOT} could benefit our task. \cite{zhu2024text2mdt} constructs a series of different COT-style prompts and responses, and they find that a set of prompt-response template referred to as COT-Gen-3 performs the best. The prompt and response template are presented in Appendix \ref{sec:appendix_prompt_templates}.

\subsection{PI-LoRA: Path-Integrated LoRA}
\label{subsec:pi_lora}

Note that the previous works on Text2MDT only considers the vanilla LoRA method for fine-tuning LLMs, neglecting the other PEFT methods or more advanced variants of LoRA. Thus, in this work, we first analyze the limitations of the current LoRA methods, and then propose a novel LoRA method to enhance the fine-tuning performance. 

\subsubsection{Analysis of Problems in Existing Methods}

We now reflect on the previous representative works on LoRA rank allocation. AdaLoRA \cite{Zhang2023AdaptiveBA} first consider re-arrage the rank distributions of LoRA modules on the LLM backbone. It achieve this objective by first initialize all the LoRA modules with a large number of ranks, and prune the less important ranks gradually along with the training procedure. AdaLoRA utilize a sensitivity based importance score \cite{Michel2019sixteenhead}, 
\begin{equation}
\text{ipt}(w) = \| w \nabla_{w} \mathcal{L} \|
\end{equation}
which measures how much the training loss will change if the LoRA parameters change. However, \cite{zhang2022platon} pointed out that this importance measure is unreliable, since it only considers how one parameter change affects the model under the hypothesis that no other parameter changes occur, and have not consider its importance under different model statuses. 

AutoLoRA \cite{zhang2024autolora} builds upon the methodology of differentable neural architecture search and bi-level optimization \cite{Liu2019DARTSDA}. It considers each LoRA rank as a neural network operation and assigns a learnable architectural parameter. Its objective is to select the best LoRA architecture, which relies on the learned architectural parameters' values as the importance scores. SoRA \cite{Ding2023SparseLA} and SaLoRA \cite{Hu2023StructureAwareLA} are similar to AutoLoRA except that the architectural parameters are learned with a normal optimization procedure on the training set \cite{Bi2020GOLDNASGO}. The LoRA ranks with higher architectural weights are kept while others are pruned. However, as pointed out by \cite{chen2020stabilizing}, the architectural parameters can not reliably reflect the quality or importance of the LoRA ranks.

To enhance the effectiveness of the LoRA scoring mechanism, we identify the key challenge as overcoming the limitations of sensitivity scores (the foundation of the AdaLoRA method). Our primary inspiration stems from Shapley value theory \cite{lundberg2017unified}. This theory models module evaluation as a cooperative game framework, treating each module as an independent participant. The Shapley value $\Phi(m)$ for a neural network module $m$ is defined by the following game-theoretic equation:
\begin{equation}
\Phi_m = \frac{1}{|\mathcal{S}_k|} \sum_{A \subseteq \mathcal{S}_k \setminus \{m\}} \Big[ \mathcal{V}(A \cup \{m\}) - \mathcal{V}(A) \Big]
\label{eq:shapley_game}
\end{equation}
where $\mathcal{S}_k$ represents the space of all possible module combinations, and $\mathcal{V}(\cdot)$ denotes the coalition utility function. This framework captures synergistic effects through multi-dimensional interaction evaluation. While computational complexity remains challenging, its theoretical foundation provides crucial insights for our research.

Our second inspiration comes from gradient path integration techniques. The integrated gradients method \cite{sundararajan2017axiomatic} measures the importance by integrating along paths:
\begin{equation}
\text{ipt}(\omega) = \omega * \int_{\alpha=0}^{1} \frac{\partial F(\alpha \omega)}{\partial \omega} d\alpha
\label{eq:integrated_grad}
\end{equation}
This method constructs an interpolation path between baseline input $\boldsymbol{x}'$ (e.g., zero vector) and target input $\boldsymbol{x}$, quantifying feature contributions through gradient integration. Essentially, it computes gradients under different parameter scaling coefficients $\alpha$, though significant computational overhead arises from multiple forward passes.

\subsubsection{PI-LoRA}

Building on the above analysis, we propose PI-LoRA to improve importance evaluation of LoRA modules in LLMs. Given any parameter $w \in \text{LoRA}_{m, l}$, loss function $\mathcal{L}$, and zero baseline, we construct the importance scoring function:
\begin{align}
s(w) &= \left| w \int_0^1 \nabla_{w} \mathcal{L}(\alpha w)  d\alpha \right| \label{eq:ig_core} \\
&\approx \frac{|w|}{K} \left\| \sum_{k=1}^{K} \nabla\mathcal{L}(\tfrac{k}{K} w ) \right\| \label{eq:trap_approx}
\end{align}
Equation \eqref{eq:trap_approx} employs the trapezoidal rule with $K > 0$ equidistant points to approximate high-dimensional integration, addressing the strong non-convexity of $\mathcal{L}$ in LLM parameter spaces.

Equation \eqref{eq:trap_approx} requires $K$ gradient computations, yielding $O(K)$ complexity. To reduce computational load, we design a stochastic sampling strategy. Assume the training process contains $P$ training steps. During the $p$-th mini-batch of fine-tuning, we uniformly sample $\alpha_p$ from $\{\frac{1}{K},...,\frac{K-1}{K}, \frac{K}{K}\}$, with single-point approximation:
\begin{equation}
\tilde{s}^{(p)}(w) =\left\|  (\alpha_p * w) * \nabla\mathcal{L}(\alpha_p * w ) \right\|
\label{eq:stoch_approx}
\end{equation}
After $P$ mini-batches, parameter importance is obtained via temporal aggregation:
\begin{equation}
\tilde{s}(w) = \frac{1}{P} \sum_{p=1}^{P} \tilde{s}^{(p)}(w).\label{eq:importance_agg}
\end{equation}
This approach reduces complexity to $O(1)$ per parameter per batch, significantly enhancing feasibility for large-scale models. Then the sensitivity score for the whole LoRA module is calculated as the average score:
\begin{equation}
\tilde{s}( \text{LoRA}_{m, l} ) = \dfrac{  \sum_{ w\in \text{LoRA}_{m, l} }  \tilde{s}(w) }{ | \text{LoRA}_{m, l} | }.
\label{eq:lora_score}
\end{equation}

\subsubsection{Algorithm Overview}

We now present the overall procedure for our PI-LoRA method, which is presented in Algorithm \ref{alg_1}. We can see that PI-LoRA is a LoRA pruning method equipped with our LoRA scoring method. We calculate the LoRA importance scores from Eq~\ref{eq:lora_score}, and then prune the LoRA parameters that receive the lowest scores.

\begin{algorithm}[h]
    \caption{PI-LoRA}
    \label{alg_1}
    \begin{algorithmic}[1]
    \REQUIRE Train data $\mathcal{D}$; the number of training steps $P$; randomly initialized LoRA modules $\text{LoRA}_{m, l}$ ($l < L$, $m \in S_{ \text{LoRA} }$); targeted number of LoRA modules $N_{\text{LoRA}}$.
    
    \FOR{$p = 1$ to $P$}
       \STATE  Sample a mini-batch data from $\mathcal{D}$;
        \STATE Sample $\alpha_{p}$ from $\{\frac{1}{K},...,\frac{K-1}{K}, \frac{K}{K}\}$; 

        \STATE Conduct a forward pass on $\text{Ba}$ and compute the gradients of the LoRA modules;

        \STATE Calculate the sensitivity scores $\tilde{s}^{(p)}( \text{LoRA}_{m, l} )$ via Eq~\ref{eq:importance_agg};

        \STATE Accumulate the average score of $\text{LoRA}_{m, l}$ for each LoRA module via Eq~\ref{eq:lora_score}.
    \ENDFOR
    
    \STATE Prune the LoRA modules that receives the lowest scores, so that remaining LoRA parameters meet the targeted number of LoRA modules $N_{\text{LoRA}}$. 
       
    \end{algorithmic}
    \end{algorithm}

\section{Experiments}
\label{sec:experiments}

\subsection{Datasets and evaluation metrics}

In this work, we mainly use the Text2MDT task for evaluation. Readers are referred to \cite{zhu2024text2mdt} for detailed introductions and dataset statistics. In order to evaluate how different models perform on the Text2MDT task, we now define the following evaluation metrics: 
\begin{itemize}
\item For triplet extraction, we follow \cite{PromptCBLUE} to adopt the triplet-level precision (Prec), recall (Rec) and F1 scores as evaluation metrics.
\item For node grouping, we define a Levenshtein ratio \cite{Navarro2001AGT} style score, NG\_LR, for this subtask. 
\item For the tree assembling subtask and also the whole Text2MDT task, we define three metrics: (a) the accuracy of decision tree extraction (Tree\_Acc); (b) the F1 score of decision paths (DP\_F1); (c) Lenvenshtein ratio of the decision tree (Tree\_LR).

\end{itemize}

\subsection{Baselines}

On the Text2MDT task, We compare our method with the current SOTA PEFT baseline methods. 

\noindent\textbf{LoRA and its variants} \ we consider the following LoRA variants as baselines: (a) the original LoRA \cite{hu2021lora} which are considered by \cite{zhu2024text2mdt}; (b) AdaLoRA \cite{Zhang2023AdaptiveBA}, which adaptively adjust the LoRA parameters among different Transformer modules. (c) AutoLoRA \cite{zhang2024autolora}, which utilize the bi-level optimization method \cite{Liu2019DARTSDA} to learn the LoRA ranks' importance scores. (d) MOELoRA \cite{liu2023moelora}, which considers each LoRA module as a mixture of single-rank LoRA experts. (e) DoRA \cite{liu2024dora}.

\noindent\textbf{Other PEFT methods} \ We also consider the most recent PEFT methods: (a) Parallel-Adapter proposed by \citet{He2021TowardsAU}; (b) Learned-Adapter \cite{Zhang2023LearnedAA}. (c) P-tuning v2 \cite{Liu2021PTuningVP}. (d) IAPT \cite{zhu2024iapt}. (e) BitFit \cite{BenZaken2021BitFitSP}. (f) (IA)$^{3}$ \cite{Liu2022FewShotPF}, which multiplies learnable vectors to the hidden states in different modules of the Transformer layer. (g) SSP \cite{Hu2022SparseSS}.

The baselines are implemented using their open-sourced codes. We only adjust the hyper-parameters related to tunable parameter numbers to fairly compare the baseline methods and our method.

\begin{table*}
\centering
\resizebox{0.96\textwidth}{!}{
\begin{tabular}{cccccccc}

\hline
 \textbf{Subtask}   &    \multicolumn{3}{c}{\textbf{Triplet extract}}    &   \textbf{Node Grouping}    &    \multicolumn{3}{c}{\textbf{Tree assembling}}       \\ 
 \textbf{Metric}   &   Prec   &  Rec   &  F1   &  
 $\text{NG\_LR}$     &  Tree\_Acc  & DP\_F1   &   Tree\_LR     \\
\hline

\multicolumn{8}{c}{\emph{LLM APIs}}        \\ 
\hline

GPT-4  &   0.783     &     0.815   &   0.798   &   0.916    &     0.672   &     0.786    &    0.893        \\
\hline
\multicolumn{8}{c}{\emph{Encoder-based methods}}        \\ 
\hline
UNIRE   &    0.913  &  0.881  &    0.896    &   - &  -  &  -  &  -     \\
TPLinker   &   0.909   & 0.878    &  0.893    &  - &  -  &  -  &  -          \\
CasRel    &    0.882   &  0.891   &  0.886     &  - &  -  &  -  &  -      \\
Sep-Biaffine   &    0.893   &  0.897   &   0.895     &      - &  -  &  -  &  -   \\ 

NG-Biaffine   &      -  &    -  &   -  &    0.962    &     - &  -  &  -      \\
NG-TableFilling    &   -    &     -   &   -   &     0.961    &    - &  -  &  -      \\ 

TreeAssemble-Biaffine   &   -   &    -   &    -    &    -    &    0.735    &    0.841   &   0.937              \\
TreeAssemble-TableFilling    &   -   &   -   &  -    &     -   &    0.741     &   0.838   &   0.933    \\

\hline
\multicolumn{8}{c}{\emph{LLM fine-tuning methods}}        \\ 
\hline

IA3    &    0.886   &    0.902    &    0.893    &    0.957    &  0.746    &   0.835    &   0.933  

\\
Bitfit     &     0.879    &   0.911   &    0.894    &    0.962    &   0.741   &  0.828   &    0.924     \\

LoRA  &  0.901   &  0.908  &   0.904    &  0.973    &   0.764   &   0.858    &   0.952     \\

AdaLoRA     &   0.903   &   0.910   &   0.906   &   0.978    &   0.761   &   0.852   &   0.948        \\

AutoLoRA    &    0.914   & 0.901    &     0.907   &   0.976   &   0.765   &  0.860  &   0.954       \\

MOELoRA     &    0.910   &   0.907    &   0.908    &   0.975    &   0.764   &   0.859   &    0.953     \\
DoRA      &  0.906   &   0.913    &     0.909   &   0.978   &   0.767   &   0.862  &    0.959          \\

PI-LoRA (ours)   &      0.911  &   0.916    &     0.913    &    0.981   &   0.772   &   0.884   &   0.967    \\

\hline

\end{tabular}}
\caption{\label{tab:main_results_1}Results for each subtask in Text2MDT. The average results in five different runs are reported. The best results are in bold. }
\end{table*}

\subsection{Experimental settings}

\noindent\textbf{LLM backbones} \quad The main experiments use the most recent open-sourced LLM,  Qwen 2.5 7B models \cite{yang2025qwen3} as the pretrained backbone model. In the ablation studies, we will also use the Baichuan 2 7B models \cite{yang2023baichuan} and GLM-4-9B-Chat\footnote{https://huggingface.co/zai-org/glm-4-9b-chat-hf}. When fine-tuning a LLM, we only consider the supervised fine-tuning (SFT) setting \cite{ouyang2022training}. After receiving a prompt or instruction, all the predictions are generated using the language modeling head (LM head). No additional prediction heads are installed for making categorical or numerical predictions. For decoding during inference, we use beam search with beam size 3.

\noindent\textbf{Implementation details for PI-LoRA} \quad For our PI-LoRA method, each LoRA module is initialized with rank $r = 16$. For our PI-LoRA method, $K$ is set to 25. And at the end of the PI-LoRA training procedure, half of the LoRA modules will be pruned. And the remaining LoRA placements will be used as the LoRA setting for re-initialization and retraining. The Adam optimizer \cite{Loshchilov2017DecoupledWD} is employed throughout all experiments. The loss objective is MSE. The learning rate is set to 1e-4, and the number of learning rate warm-up steps is 100. The batch size is set to 32, with the help of gradient accumulation technique. We run validation on the valid set after each epoch. If the validation loss does not drop for 5 epochs, then the training will stop. The gradient checkpoints with the lowest validation loss will be used to make predictions on the test set. 

During the final fine-tuning stage, all the LoRA modules are randomly initialized according to the allocation setting delivered by the previous stage. And training hyper-parameters are set to be the same with the previous stage. In every 200 steps, the model is evaluated on the dev set. Patience is set to 10, that is, if the model does not achieve a lower development set loss for 10 evaluation runs, the training stops. The best checkpoint on the dev set is used to run predictions on the test set.

\subsection{Main results}

\noindent \textbf{Results on the sub-tasks} \quad In this setup, We compare our proposed method with baseline PEFT methods by employing these methods in fine-tuning on the Text2MDT task. The experimental results are presented in Table \ref{tab:main_results_1}. We present the encoder-based methods from \cite{zhu2024text2mdt} as comparison. Table \ref{tab:main_results_1} reveals that our PI-LoRA method outperforms the baseline methods across all seven tasks, with comparable tunable parameters. In particular, PI-LoRA outperforms the previous SOTA LoRA-based baselines like AdaLoRA, AutoLoRA, DoRA, and MOELoRA with comparable or less tunable parameters. These results demonstrate that our method excels at downstream task adaptation of large language models. In addition, we can see that our PI-LoRA method outperforms all the encoder-based methods.

\noindent \textbf{Results on the end2end framework} \quad In this set of experiments, we adopt the end2end framework and fine-tune the LLM to complete the whole MDT generation process upon the corresponding prompt. The results are detailed in Table \ref{tab:main_results_2}. Consistent with earlier findings (Table \ref{tab:main_results_1}), PI-LoRA achieves superior performance compared to the baseline methods across all benchmarks. These results underscore PI-LoRA’s efficacy in improving instruction-tuning performance for LLMs, highlighting its potential as a robust alternative to existing parameter-efficient adaptation strategies.

\begin{table}
\centering
\resizebox{0.48\textwidth}{!}{
\begin{tabular}{cccc}
\hline
 \textbf{Method}   &     Tree\_Acc  & DP\_F1   &   Tree\_ER       \\
\hline

LoRA   &   0.510  &   0.646   &   0.911     \\

AdaLoRA     &    0.510   &   0.651   &     0.907    \\

AutoLoRA     &   0.510   &   0.648   &     0.914            \\

MOELoRA     &   0.520   &   0.657      &    0.917     \\
DoRA     &   0.520   &   0.660      &    0.923                   \\

PI-LoRA (ours)   &    0.550   &   0.679      &    0.936           \\                                                                                     \hline
\end{tabular}}
\caption{\label{tab:main_results_2}Overall results of the end2end methods using different PEFT methods. The average results in five different runs are reported. The best results are in bold.  }

\end{table}

\subsection{Ablation studies and further analysis}

\noindent \textbf{Sensitivity analysis on the $K$ parameter} \quad In our main experiments (Table \ref{tab:main_results_1} and \ref{tab:main_results_2}), we set the $K$ parameter to 25. Now we change its value to \{1, 5, 10, 15, 50, 100\}, and investigate how it affects the fine-tuning performance. The experimental results are presented in Table \ref{tab:ablation_on_k}. The results show that $K > 10$ results in reasonable performances, and our PI-LoRA method is robust to this parameter. However, low $K$ results in sub-optimal performance. This is natural: low $K$ values means that our method can not properly approximate the path integral's process, making our method to reduce to the AdaLoRA method.

\begin{table}
\centering
\resizebox{0.46\textwidth}{!}{
\begin{tabular}{cccc}
\hline
 \textbf{Value of $k$}   &     Tree\_Acc  & DP\_F1   &   Tree\_ER       \\
\hline

$K=1$   &   0.510  &   0.652   &   0.906     \\

$K=5$   &    0.520   &   0.664      &  0.923      \\
$K=10$   &    0.530   &   0.672      &  0.928      \\
$K=15$   &    0.540   &   0.678      &  0.934      \\
$K=25$   &    0.550   &   0.679      &    0.936           \\             
$K=50$   &    0.550   &   0.677      &    0.935           \\ 
$K=100$   &    0.550   &   0.679      &    0.935          \\ 

\hline
\end{tabular}}
\caption{\label{tab:ablation_on_k}Results of different values for the $K$ parameter.  }

\end{table}

\noindent\textbf{On the stability of our scoring method} \quad On a given LLM backbone, we need to investigate whether our LoRA scoring and pruning is stable since it involves random sampling. We run the whole LoRA scoring procedure under 3 different random seeds, and compare the LoRA importance scores obtained in each run. We calculate the similarities between these three sets of scores pairwise. The similarity score is measured using Spearman rank correlation. Note that these three results are not included in the evaluation of the previous experiments. We present the pairwise correlations in Table \ref{tab:results_different_random_seeds}. From the results, we can see that the importance scores of the LoRA ranks obtained under different random seeds have very high correlations, indicating that the LoRA scores obtained by our method is stable with respect to random seeds.

\begin{table}
\centering
\resizebox{0.38\textwidth}{!}{
\begin{tabular}{c|ccc}
\hline
         &  Seed 1   &   Seed 2    &   Seed 3     \\
\hline
Seed 1   &    1.00   &   0.94     &   0.93 \\
Seed 2   &      -     &   1.00      &   0.92    \\
Seed 3   &     -      &      -       &   1.00   \\

\hline
\end{tabular}}
\caption{The pairwise similarity scores for the LoRA ranks' importance estimations obtained under three random seeds. }
\label{tab:results_different_random_seeds}
\end{table}

\noindent \textbf{Ablation studies of the LLM backbone} \quad Our main experiments are conducted on the Qwen 2.5 7B model. To demonstrate the wide applicability of our method, we now conduct experiments on the Baichuan 2 7B and GLM-4-9B-Chat. The results are reported in Table \ref{tab:results_different_backbones}. We can see that on these two backbones, our method can also outperform the baseline methods.

\begin{table}
\centering
\resizebox{0.44\textwidth}{!}{
\begin{tabular}{cccc}
\hline
 \textbf{Method}   &     Tree\_Acc  & DP\_F1   &   Tree\_ER       \\
\hline 
\multicolumn{4}{c}{\textbf{\emph{Results for Baichuan 2 7B }}}  \\
\hline 

LoRA    &    0.490  &   0.632  &   0.898  \\
AdaLoRA   &   0.490  &   0.634   &   0.903   \\
Ours     &  0.51   &   0.645   &   0.909    \\

\hline 
\multicolumn{4}{c}{\textbf{\emph{Results for GLM-4-9B-Chat }}}  \\
\hline 

LoRA  &  0.540   &   0.678   &    0.936 \\
AdaLoRA  &  0.540   &  0.676  &    0.935   \\
Ours     &   0.560   &   0.688   &   0.942   \\

\hline
\end{tabular}}
\caption{\label{tab:results_different_backbones}Results for different PEFT methods, when the backbone LLMs are Baichuan 2 7B and GLM-4-9B-Chat. }

\end{table}

\section{Conclusion}

In this paper, we presented FT-MDT, a novel framework for extracting medical decision trees from clinical texts through a novel low-rank adaptation method. The proposed approach addresses Text2MDT, the task of of building clinical decision support systems by automating the extraction of medical decision trees from clinical guidelines and textbooks, eliminating the need for time-consuming manual annotation. The key innovation of our work lies in the development of PI-LoRA (Path-Integrated LoRA), which overcomes the limitations of existing rank allocation methods in LoRA by leveraging Shapley value theory and gradient path integration. Unlike previous approaches that rely on unreliable sensitivity scores, PI-LoRA provides a more effective and theoretically grounded mechanism for module-specific rank allocation, significantly improving the efficiency and performance of the model. Our experimental results demonstrate that the PI-LoRA method achieves superior performance compared to previous approaches. This work not only advances the state-of-the-art in medical decision tree extraction but also provides a valuable framework for LoRA-based LLM fine-tuning.

Future work will explore extending this framework to handle more complex medical decision-making scenarios and investigate its applicability across diverse medical domains with varying levels of structural complexity. The principles underlying PI-LoRA may also inspire more effective parameter-efficient fine-tuning methods for other natural language processing tasks involving hierarchical structure extraction.

\section*{Limitations}

Despite the promising results achieved by our PI-LoRA approach for medical decision tree extraction, several limitations warrant consideration. First, while our method significantly reduces the need for manual annotation, it remains sensitive to the quality and structure of the input medical texts. Clinical guidelines with non-standard formatting, ambiguous phrasing, or complex multi-step decision processes often lead to incomplete or inaccurate tree structures, particularly when the text contains multiple concurrent treatment paths that are not clearly delineated.

Second, the prompt engineering approach, while effective for standard cases, faces challenges with rare medical conditions or emerging treatment protocols that fall outside the training distribution of the LLM. The current triplet relationship categories (e.g., "clinical manifestations," "therapeutic drugs") may not adequately capture specialized medical knowledge in rapidly evolving fields, potentially requiring domain-specific customization for optimal performance.

Third, the PI-LoRA method, while computationally efficient compared to full fine-tuning, still requires careful tuning of the path integration parameters for different medical domains. The current implementation assumes a relatively uniform structure across medical texts, which may not hold for highly specialized clinical guidelines that deviate from the standard binary decision tree format.

Fourth, our evaluation primarily focuses on structural accuracy of the extracted decision trees, but does not fully address the clinical validity of the resulting decision logic. A tree that is structurally correct may still contain medically questionable recommendations due to limitations in the LLM's knowledge base or the prompt design.

Finally, the current implementation is limited to English-language medical texts, which restricts its immediate applicability in non-English speaking healthcare settings without significant additional adaptation. Future work should address these limitations to broaden the method's practical utility across diverse medical contexts and languages.

\section*{Ethics Statement}

This research involves the extraction of medical decision trees from clinical texts using large language models. We have carefully considered the ethical implications of our work in the healthcare domain and have implemented the following measures:

Data Privacy: All medical texts used in this study were publicly available clinical guidelines or synthetic data generated for research purposes. No patient-identifiable information was included in any of our datasets or models. All data processing and model training complied with institutional privacy policies.

Bias Mitigation: We acknowledge that medical data may contain biases that could affect clinical decision-making. To address this, we implemented a multi-step verification process involving medical professionals to review the extracted decision trees for potential biases in treatment recommendations. We recognize that our current model may not capture all nuances of medical decision-making and recommend future work with more diverse medical data sources.

Safety and Clinical Use: The extracted decision trees are intended to support, not replace, clinical decision-making. We explicitly state that all clinical decisions must be made by qualified healthcare professionals, and our system should not be used as a standalone diagnostic or treatment tool. We emphasize the importance of human oversight in any clinical implementation.

Transparency: We provide a comprehensive description of our methodology, model architecture, and limitations to enable peer review, replication, and critical evaluation of our work. The prompt templates, evaluation metrics, and implementation details are fully documented for transparency.

Potential Misuse: We recognize that this technology could be misused to automate clinical decisions without appropriate oversight. Therefore, we recommend that our system should only be deployed in clinical settings with appropriate regulatory approvals and under the supervision of qualified medical professionals.

Social Impact: While we believe this research has the potential to improve clinical decision support systems and enhance healthcare efficiency, we acknowledge the need for ongoing discussion about the societal implications of increased automation in healthcare, including potential impacts on healthcare workforce dynamics. We commit to ongoing ethical review of our work as the technology develops and is implemented in real-world clinical settings.

\bibliography{custom}
\bibliographystyle{acl_natbib}

\appendix

\section{Prompt templates and response formats for the pipeline framework}
\label{sec:appendix_prompt_templates}

\subsection{The triplet extraction subtask}

In the triplet extraction task asks a language model to predict a series of triplets from the given text. A triplet includes the head entity mention, tail entity mention, and the relation between them. For the triplet extraction sub-task, the template for the input prompt is: 
\begin{lstlisting} 
Please extract triplets based on the following medical guideline text:

[Text]

Instruction: Extract the triplet used to describe diagnosis and treatment knowledge or clinical information as the content of the condition/decision node. The triplet relationship defines a total of 6 categories: "clinical manifestations", "therapeutic drugs", "usage and dosage", "Treatment plan", "Prohibited drugs", "Basic situation"
\end{lstlisting} 
the response should be formated as follows:
\begin{lstlisting} 
The triples in the given guideline text are as follows:

[triplets]
\end{lstlisting} 
The special token [Text] denotes the input text, and [triplets] denotes a list of triplets.

\subsection{The node grouping subtask}

In the node grouping task, we asks a language model to predict which triplets form a node, and which logical relation the node has. For the node grouping sub-task, the template for the input prompt is: 
\begin{lstlisting} 
Please combine these triples into several nodes based on the following medical guideline text and the triplet information extracted from it, and indicate the logical relationship of the triplets within this node:

Medical guideline text:

{Text}

The triples in the given guideline text are as follows:

{triplets}

Note: If several triples form a node, it means that there is an and or or logical relationship between these triples. If a triple does not have an and or or relationship with other triples, it means that the triple needs to become a node independently.
\end{lstlisting} 
the response should be formated as follows:
\begin{lstlisting} 
Based on the given guideline text and its triplet information, the nodes of the decision tree are composed as follows:

The following triples constitute a node of the decision tree: [triplets]. The logical relationship of this node is: [logical_rel]

The following triples constitute a node of the decision tree: [triplets]. The logical relationship of this node is: [logical_rel]
\end{lstlisting} 
The special token [Text] denotes the input text, and [triplets] denotes the list of extracted triplets, and [node] denotes the contents of the node.

\subsection{The tree assembling subtask}

In the tree assembling task, given the results of the node grouping step, we ask the language model to generate the whole decision tree. For the tree assembling sub-task, the template for the input prompt is: 
\begin{lstlisting} 
Please form a decision tree based on the following medical guideline text and the node information extracted from it:

Medical guideline text:

[Text]

The nodes in a given guideline text are composed as follows:

[nodes]

Note: (1) The diagnosis and treatment decision tree is a binary tree composed of conditional nodes and decision nodes. It aims to express guideline text through concise structured information. It requires not only to dig out the core entities and relationships in the text, but also to carry out this information. They are connected in series to form a complete decision-making process; (2) In the diagnosis and treatment decision-making binary tree, non-leaf nodes are condition nodes and leaf nodes are decision nodes. For the condition node, when the condition judgment result is "yes", it will go to the left child node for the next judgment or decision. When the condition judgment result is "no", it will go to the right child node for the next judgment or decision. (3) The output of each node is a dict, containing three fields: (3a) "role", which is the node role type; (3b) "triples", which is a list of triples; (3c) "logical_rel", which represents the node logical relationship. (4) The entire diagnosis and treatment decision tree is arranged into a list using the breadth-first strategy.
\end{lstlisting} 
the response should be formated as follows:
\begin{lstlisting} 
The diagnosis and treatment decision tree extracted based on the given guideline text is as follows:

Node [node_idx]: role=[role]; logical_rel=[logical_rel]; triplets=[triplets]

Node [node_idx]: role=[role]; logical_rel=[logical_rel]; triplets=[triplets]

Node [node_idx]: role=[role]; logical_rel=[logical_rel]; triplets=[triplets]
\end{lstlisting} 
The special token [Text] denotes the input text, and [nodes] denotes the list of nodes from the previous subtask. In the response, [node\_idx] denotes the index of a node, [triplets] denotes the list of extracted triplets in a node, [logical\_rel] denotes the logical relation of the node, and [role] denotes the role label of the node.

\subsection{COT-Gen-3}

Following \cite{zhu2024text2mdt}, we consider the following template for the COT-Gen-3 method:
\begin{lstlisting} 
Please generate a diagnosis and treatment decision tree for the following medical guideline text:

[Text]

Task description: 
(1) Based on the given medical guideline text, create a binary tree, including conditional nodes and decision nodes, to succinctly display the guideline content while capturing core entities and relationships;  
(2) Conditional nodes are used for judgment, based on the results point to the left or right child node to make the next decision.  
(3) The output of each node is a dict, containing three fields:  
    - (3a) "role" indicates the node type, which can be a condition node ("C") or a decision node ("D");  
    - (3b) "triples" is a list of triples describing diagnosis and treatment knowledge or clinical information, including "clinical manifestations", "therapeutic drugs", "usage and dosage", "treatment plan", "Prohibited drugs", "Basic situation" six types of relationships;  
    - (3c) "logical_rel" represents the logical relationship between multiple triples (the values are and, or, null, when there is only one triple. The logical relationship is null when it is a tuple).  
(4) The finally generated diagnosis and treatment decision tree is arranged into a list according to the breadth-first strategy.)

Instructions for the generation steps: Please complete the generation of the decision tree step by step. (a) First extract triples from the above text; (b) and then generate a complete decision tree.
\end{lstlisting} 
where [Text] is a piece of medical text. According to \cite{zhu2024text2mdt}, the response should be formated as follows:
\begin{lstlisting} 
The triples in the given guideline text are as follows:

[triplets]

The diagnosis and treatment decision tree extracted based on the given guideline text is as follows:

Node [node_idx]: role=[role]; logical_rel=[logical_rel]; triples=[triplets]  
Node [node_idx]: role=[role]; logical_rel=[logical_rel]; triples=[triplets]  
Node [node_idx]: role=[role]; logical_rel=[logical_rel]; triples=[triplets]  
\end{lstlisting} 
where [node\_idx] represents the node id, [role] means the role of the node, [logical\_rel] stands for the logical relation in the node, and [triplets] expresses the contents in the node. From the above prompt template and response, we can see that COT-Gen-3 asks the LM to extract triplets and then generate the whole MDT.

\end{CJK*}

\end{document}